\UseRawInputEncoding
\documentclass[lettersize,journal]{IEEEtran}
\pdfoutput=1
\usepackage{amsmath,amsfonts}
\usepackage{algorithmic}
\usepackage{algorithm}
\usepackage{array}
\usepackage{textcomp}
\usepackage{stfloats}
\usepackage{url}
\usepackage{verbatim}
\usepackage{graphicx}
\usepackage{cite}
\usepackage{bm}
\usepackage{subfigure}
\usepackage{bbding}
\usepackage[justification=centering]{caption}

\begin{document}

\title{Rotation center identification based on geometric relationships for rotary motion deblurring}

\author{Jinhui Qin, Yong Ma, Jun Huang, Fan Fan, You Du
	\thanks{This work was supported in part by the National Natural Science Foundation of China under Grant 62075169 and U23B2050, and in part by the Industry-University-Research Cooperation Program of Zhuhai under Grant 2220004002828. \textit{(Corresponding authors: Jun Huang.)}}
	\thanks{Jinhui Qin, Yong Ma, Jun Huang, Fan Fan, and You Du are with the Electronic Information School, Wuhan University, Wuhan, 430072, China (e-mail: Jinhui\_Qin@163.com; mayong@whu.edu.cn; junhwong@whu.edu.cn; fanfan@whu.edu.cn; dy0815@whu.edu.cn).}}



\maketitle

\begin{abstract}
Non-blind rotary motion deblurring (RMD) aims to recover the latent clear image from a rotary motion blurred (RMB) image. The rotation center is a crucial input parameter in non-blind RMD methods. Existing methods directly estimate the rotation center from the RMB image. However they always suffer significant errors, and the performance of RMD is limited. For the assembled imaging systems, the position of the rotation center remains fixed. Leveraging this prior knowledge, we propose a geometric-based method for rotation center identification and analyze its error range. Furthermore, we construct a RMB imaging system. The experiment demonstrates that our method achieves less than 1-pixel error along a single axis (x-axis or y-axis). We utilize the constructed imaging system to capture real RMB images, and experimental results show that our method can help existing RMD approaches yield better RMD images.
\end{abstract}


\section{Introduction}\label{Sec:introduction}

\begin{figure}
	\centering
	\subfigure[The blurred image.]{
		\includegraphics[scale=0.42]{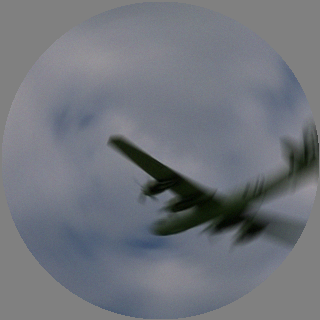}
		\label{Fig:center_error_blur}}
	\subfigure[The result with $\delta p=0$.]{
		\includegraphics[scale=0.42]{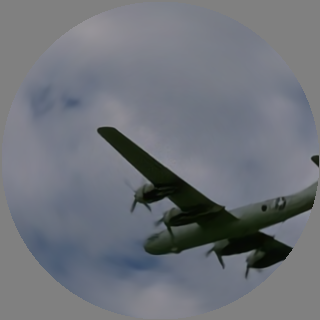}
		\label{Fig:center_error0}}
	\subfigure[The result with $\delta p=1$.]{
		\includegraphics[scale=0.42]{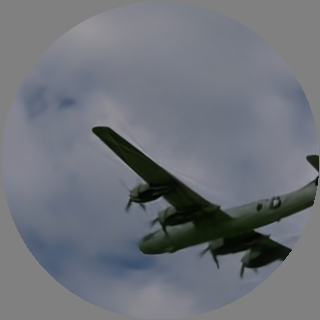}
		\label{Fig:center_error1}}
	\subfigure[The result with $\delta p=10$.]{
		\includegraphics[scale=0.42]{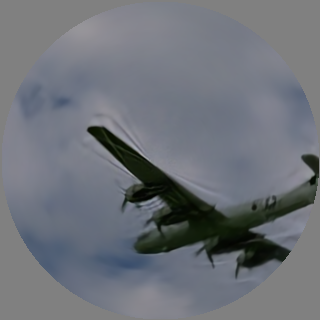}
		\label{Fig:center_error10}}
	\caption{The RMD examples with different rotation center errors $\delta p$ along the x-axis. The blurred image is degraded by additive white Gaussian noise with strength $\sigma=1\%$.}
	\label{Fig:center_error}
\end{figure}

\begin{figure}
	\centering
	\includegraphics[scale=0.48]{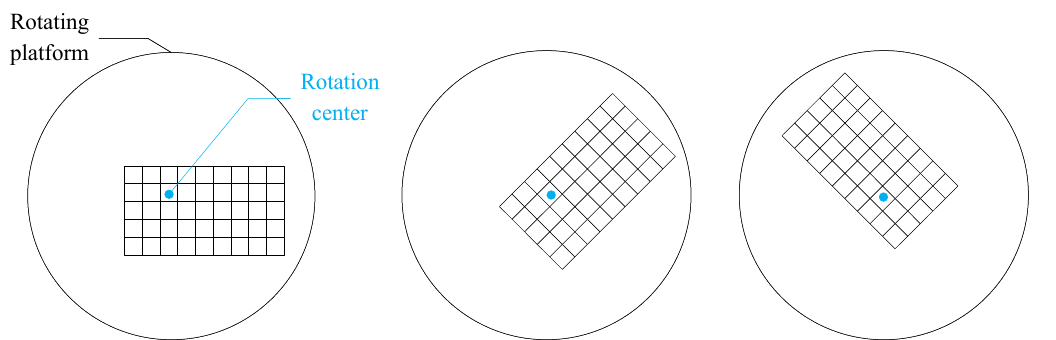}
	\caption{The position of the rotation center remains fixed under different rotation angles.}
	\label{Fig:rotation_center}
\end{figure}

Because the camera rotates with the rotating platform during exposure, captured images suffer from rotary motion blurring (RMB), which widely exists in the aerospace industry and military fields \cite{krouglova2022restoration,qiu2020end}. For example, due to the lack of attitude control systems, the satellite will suffer from an uncontrolled spin, and the camera mounted in it captures RMB images. To reconstruct details and textures removed by RMB and improve performance of high-level vision tasks, such as object recognition \cite{wang2022review}, detection \cite{cong2022weakly,zhang2023knowledge}, and tracking \cite{hu2023siammask,zheng2022leveraging}, rotary motion deblurring (RMD) becomes a crucial preprocessing procedure.

Since rotary parameters, including rotation center and blur angle, can be estimated by the RMB image, researchers regard RMD as a special non-blind deblurring \cite{qin2023progressive}. Existing methods for RMD can be divided into three categories: the model-based methods, the learning-based methods, and the progressive framework. Because the integration path of RMB is a circle, the model-based methods decompose a two-dimensional (2D) RMB image into a set of one-dimensional (1D) blurred sequences along each circle. Therefore, these blurred sequences satisfy the convolution process, and existing deblurring theories and methods can be applied to estimate latent sharp sequences. Supposing the piecewise smoothness of the natural images, Hong \textit{et al.} \cite{hong2003fast} propose the second-order difference prior (SDP) to constrain latent sharp sequences. Yuan \textit{et al.} \cite{yuan2008robust} present the modified-Wiener filtering (M-Wiener), which filters the ill-conditioned frequency components in the blur kernel to avoid overregularization. To restore strong edges and suppress noise, Wang \textit{et al.} \cite{xinchun2019image} develop the sparse adaptive prior (SAP), in which priors are estimated from blurred sequences. After being obtained from deblurring methods, latent sharp sequences are merged into the final deblurred image. The learning-based methods straight learn a mapping from the RMB image to the deblurred image. As far as we know, Qiu \textit{et al.} \cite{qiu2020end} first introduces deep learning for RMD based on generative adversarial network (GAN), marked RMD-GAN. To allow the network to perceive different blur extents, the rotary blur field is presented and fed into the generator. Furthermore, they develop a triple-scales discriminator, which effectively fuses the local and global contents. Meanwhile, since RMD is a special type of spatially variant blur \cite{hong2003fast}, other learning-based deblurring approaches designed for spatially variant blur also can be applied to RMD, such as \cite{zhang2023mc,li2023image,kong2023efficient,wang2022uformer,chen2022simple}. However, they do not utilize the knowledge on RMB image and blur kernel \cite{quan2021nonblind}. The progressive framework for RMD (RMD-PF) first combines the 1D convolution process and 2D information of the natural image and has achieved impressive visual effects \cite{qin2023progressive}. RMD-PF consists of a coarse deblurring stage and a refinement stage. In the first stage, a novel deconvolution model is constructed to balance noise suppression and details recovery. In the second stage, to reduce the ringing artifacts in the first stage, they design a triple-scale deformable module, which can adaptively adjust spatial sampling locations.

The rotation center plays a crucial role in non-blind RMD algorithms. Unlike regular motion blur, RMB images only satisfy the convolution process along each circle. Therefore, existing non-blind RMD methods need to decompose the 2D RMB image into a series of 1D blurred sequences along circles. When the rotation center is not accurate, all 1D blurred sequences deviate from the convolution process, resulting in artifacts. We use RMD-PF to evaluate the robustness of the RMD algorithm for the rotation center. As illustrated in Fig. \ref{Fig:center_error0}, when an accurate rotation center is given, this method effectively reduces RMB. However, artifacts appear in the restored image as the rotation center with errors is set. Furthermore, the larger the error, the more severe the artifacts become, as shown in Fig. \ref{Fig:center_error1} and \ref{Fig:center_error10}. Therefore, for better RMD results, a more accurate rotation center is essential. The existing approaches for rotation center determination are algorithm-based. When the rotation center is accurate, the blur extent of the 1D blurred sequence is proportional to the radius. Within a candidate region, Hong \textit{et al.} \cite{hong2003fast} find the point that satisfies this principle and is selected as the rotation center. Isoe \textit{et al.} \cite{isoe2012blind} proposed a method based on the Hough transform. They convert the RMB image into a binary image by the Laplace operator and apply the Hough transform to this binary image. The point with the highest score is the rotation center. The elements in the 1D blurred sequences, which are decomposed from the 2D RMB image, have a strong correlation. Yuan \textit{et al.} \cite{yuan2008robust} calculate this correlation in the frequency to estimate the rotation center. However, these methods have large errors and limit the performance of RMD.

The rotation center in the pixel coordinate system is consistent during the rotating process since the relative positions between the camera and the rotating platform are fixed. As illustrated in Fig. \ref{Fig:rotation_center}, the coordinate position of the rotation center remains (2,1) under different rotation angles. Therefore, for an assembled imaging system, once the rotation center is determined, the subsequent non-blind RMD or other high-level vision tasks can directly utilize this rotation center as input. In this paper, we propose a determination method for the rotation center based on geometric relationships. Furthermore, we demonstrated that the proposed method has a theoretical error of less than 0.5 pixels along a single axis (x-axis or y-axis). Finally, we constructed a RMB imaging system, in which we employed the proposed method to determine the rotation center. The experiments show that the proposed method has a single-axis error of less than 1 pixel in real-world scenarios. Due to the resonance or vibration of the step motor \cite{vernezi2021vibration,gundogdu2021narma}, the real-world rotation center error is greater than the theoretical one. The contributions of this paper are as follows:

\begin{itemize}

	\item[1)] For the first time, we propose a method for determining the rotation center in an assembled RMB imaging system. The theoretical error along a single axis is less than 0.5 pixels.

	\item[2)] We construct an RMB imaging system and demonstrate that the error of our method along a single axis (x-axis or y-axis) is less than 1 pixel in real-world scenarios.
	
\end{itemize}

\section{Methodology}\label{Sec:Methodology}
\subsection{Rotation Center Identification}\label{Sec:Identification}

\begin{figure}
	\centering
	\subfigure[]{
		\includegraphics[scale=0.22]{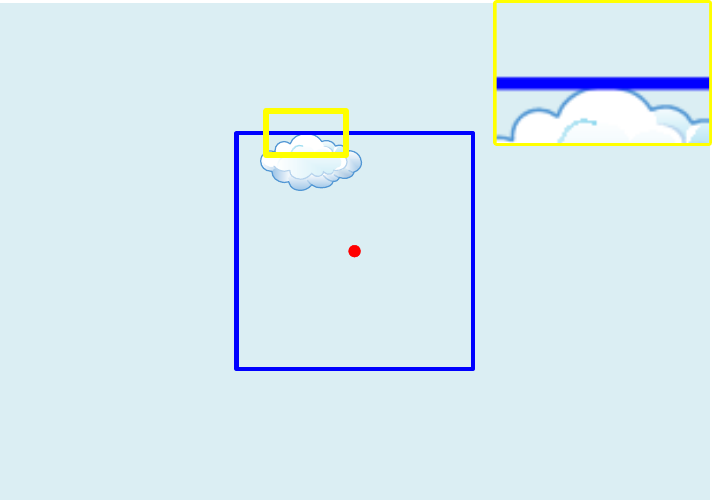}
		\label{Fig:identification_a}}\hspace{-2.5mm}
	\subfigure[]{
		\includegraphics[scale=0.22]{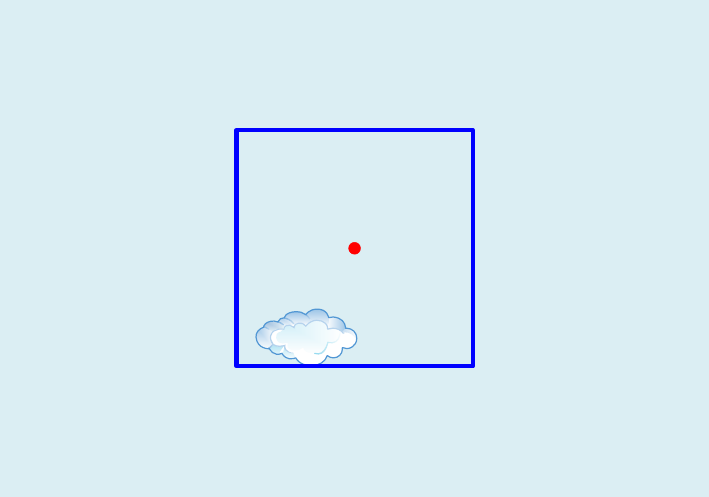}
		\label{Fig:identification_b}}\hspace{-2.5mm}
	\subfigure[]{
		\includegraphics[scale=0.22]{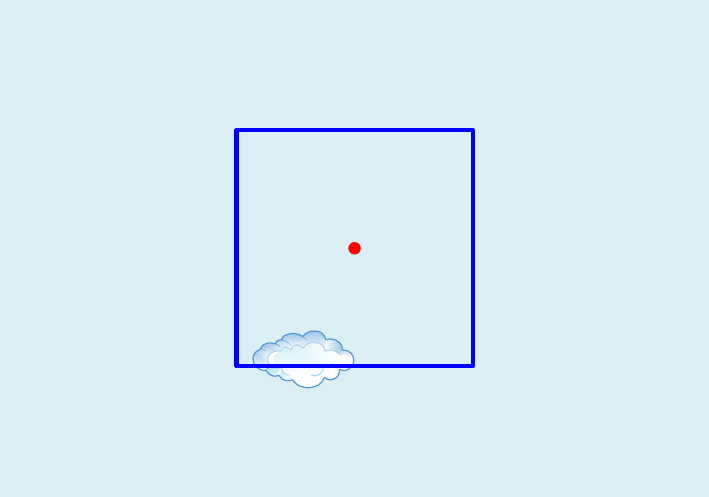}
		\label{Fig:identification_c}}
	\caption{The identification process along the y-axis. The tangent area is zoomed in, as shown in the yellow box of (a).}
	\label{Fig:identification_process}
\end{figure}

In practice, since the camera and rotating platform is fixed together, the rotation center are immutably constant once the camera is assembled. Therefore, we consider how to determine the rotation center in the assembled RMB imaging system.

This paper proposes a method for determining the rotation center based on geometric relationships. Here, we take the identification along the y-axis as an example to illustrate the principle. In the pixel coordinate system, a red point is randomly selected as a candidate rotation center, and a blue reference box is drawn centered on this point, as shown in Fig. \ref{Fig:identification_a}. During the rotation process, the positions of the red point and the blue reference box remain unchanged in the pixel coordinate system. Assuming an object which is is tangent to the top of the reference box, for example, the 'cloud' in Fig. \ref{Fig:identification_a}. Then, the rotation platform is rotated by 180°. If the red point is the ideal rotation center, the 'cloud' should be tangent to the bottom of the reference box after rotation, as shown in Fig. \ref{Fig:identification_b}; otherwise, it will not be tangent, as depicted in Fig \ref{Fig:identification_c}. Additionally, even if there is an error in the rotation center along the x-axis, it will not break this geometric relationship along the y-axis. Consequently, the identification along the x-axis and y-axis directions is independent.

According to the above principle, the rotation center identification along the y-axis is composed of the following steps:

1) Randomly select a candidate rotation center and color it red. Draw a blue reference box centered on this red point. Choose an object in the field of view, for example, the 'Cloud' in Fig. \ref{Fig:identification_a};

2) Move the camera until 'Cloud' is tangent to the blue reference box in the y-axis, as shown in Fig. \ref{Fig:identification_a}. For better observation, the tangency point is zoomed in, as illustrated in the yellow box;

3) Rotate the rotating platform 180°. If the rotated 'Cloud' is tangent to the reference box, as illustrated in Fig. \ref{Fig:identification_b}, stop the identification process; Otherwise, adjust the red point along the y-axis and return to step 2).

The identification along the x-axis is similar, so we won't go into details. Once the identifications along the y- and x-axis are done, the position of the red point is identified rotation center. Next, we analyze the error of the proposed method.

\subsection{The Error Analysis for Rotation Center}\label{Sec:Error_Analysis}

\begin{figure}
	\centering
	\includegraphics[scale=0.60]{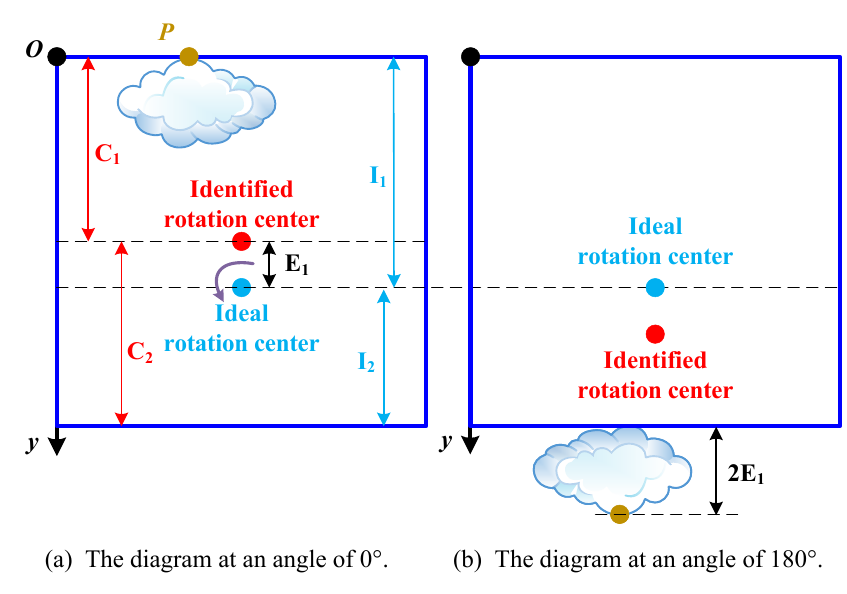}
	\caption{The diagram of rotation center error analysis.}
	\label{Fig:identification_error}
\end{figure}

The error of our identification method mainly depends on the 'tangency' between the object and the reference box: when the object differs from the reference box by one pixel, it is easy to distinguish with the naked eye; when the difference is less than one pixel, the tangency becomes ambiguous and challenging to make out.

Take the y-axis as an example, we analyze the identification error. During the whole rotation process, the position of the reference box is known and remains unchanged in the pixel coordinate system. For convenience, we establish the y-axis, whose origin is the upper left corner of the reference box. As depicted in Fig. \ref{Fig:identification_error}, the red and blue dots are identified and ideal rotation centers, respectively. The distance between the red point and the blue point is called $\rm{E}_{1}$. $\rm{C}_{1}$ and $\rm{C}_{2}$ are the distance between the red dot and upper and lower edges of the reference box, respectively. We denote the distance between the blue dot and upper and lower edges of the reference box as $\rm{I}_{1}$ and $\rm{I}_{2}$, respectively. Select an object $P$ which is tangent to the reference box. When the angle is 180°, the coordinate of point $P$ is $2\rm{I}_{1}$. The distance between $P$ and lower edges of the reference box is $2\rm{I}_{1}-(\rm{C}_{1}+\rm{C}_{2})$ = $2(\rm{C}_{1}+\rm{E}_{1})-2\rm{C}_{1}$ = $2\rm{E}_{1}$.

As mentioned above, if $\rm2{E}_{1}$ is greater than or equal to one pixel during the identification process, the observer can easily know the object is not tangent to the reference box, and re-select the next possible rotation center. Therefore:

\begin{equation}
	\label{Eq:identification_error}
	\rm2{E}_{1}<1.
\end{equation}

It indicates the identification error for the rotation center along one axis is less than 0.5 pixels.

\section{Experiments and Results}\label{Sec:Experiments and results}

\begin{figure}
	\centering
	\includegraphics[scale=0.18]{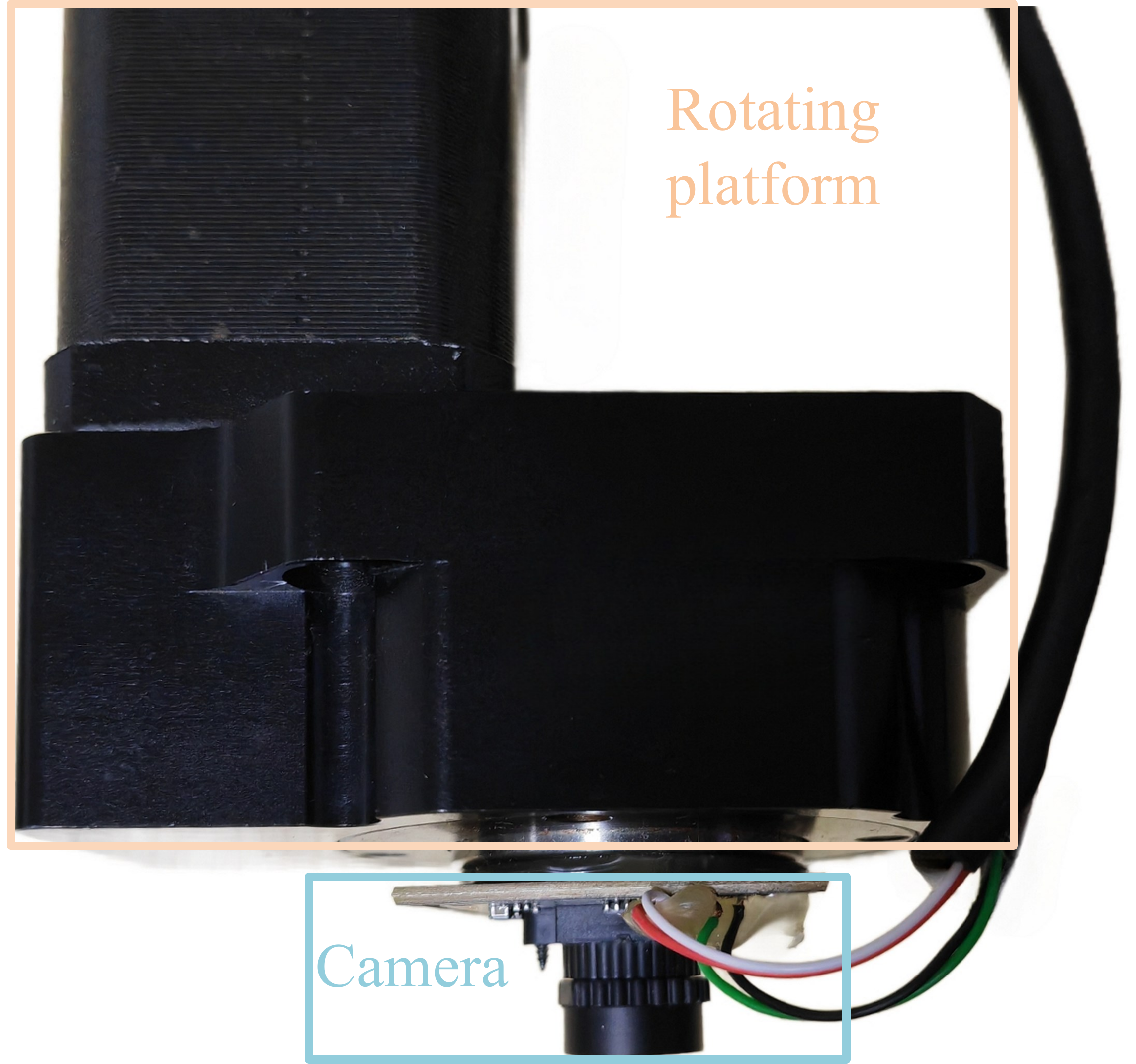}
	\caption{The top view of the RMB imaging system.}
	\label{Fig:imaging_sys}
\end{figure}

To verify the effectiveness of the proposed method, we establish a RMB imaging system. This paper utilizes a stepper motor, which rotates at a fixed angle when receiving a pulse signal, as the rotating platform. To avoid the rolling shutter effect caused by rapid motion \cite{liang2008analysis}, we use the global shutter (GS) camera. As shown in Fig. \ref{Fig:imaging_sys}, the camera is mounted on the rotating platform.

\subsection{The Results of the Rotation Center identification}

In this section, we follow the steps in Section \ref{Sec:Identification} to identify the rotation center of our constructed RMB imaging system. For better observation, we strongly recommend readers to watch the videos\footnote[1]{\url{https://pan.baidu.com/s/1MmOeeOnY7a9GYTjcOXOhwQ?pwd=s74c}} of the identification process.

\begin{figure*}
	\centering
	\subfigure[The identification process along the y-axis.]{
		\includegraphics[scale=0.185]{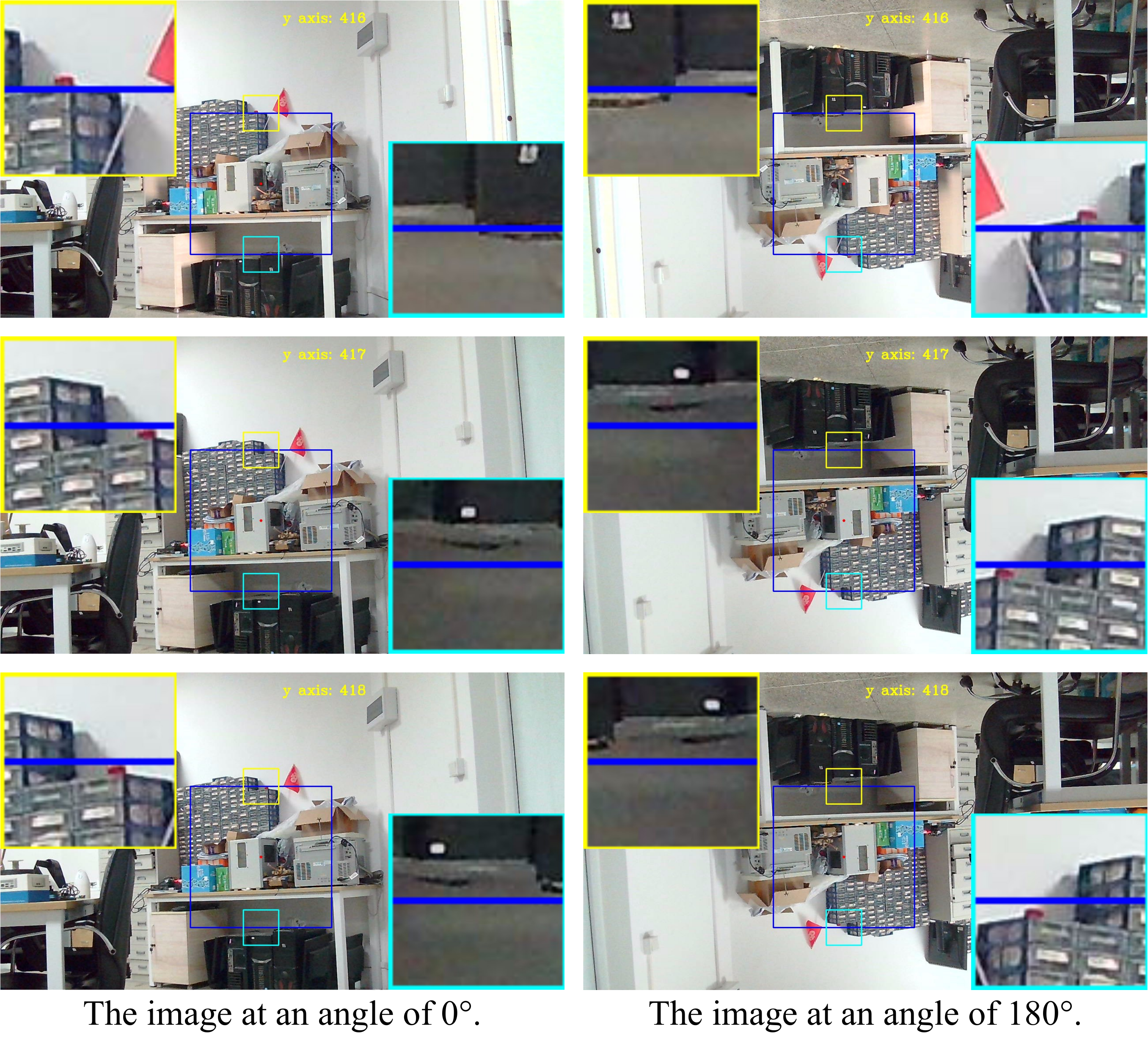}
		\label{Fig:y_aixs}}\hspace{-2mm}
	\subfigure[The identification process along the x-axis.]{
		\includegraphics[scale=0.185]{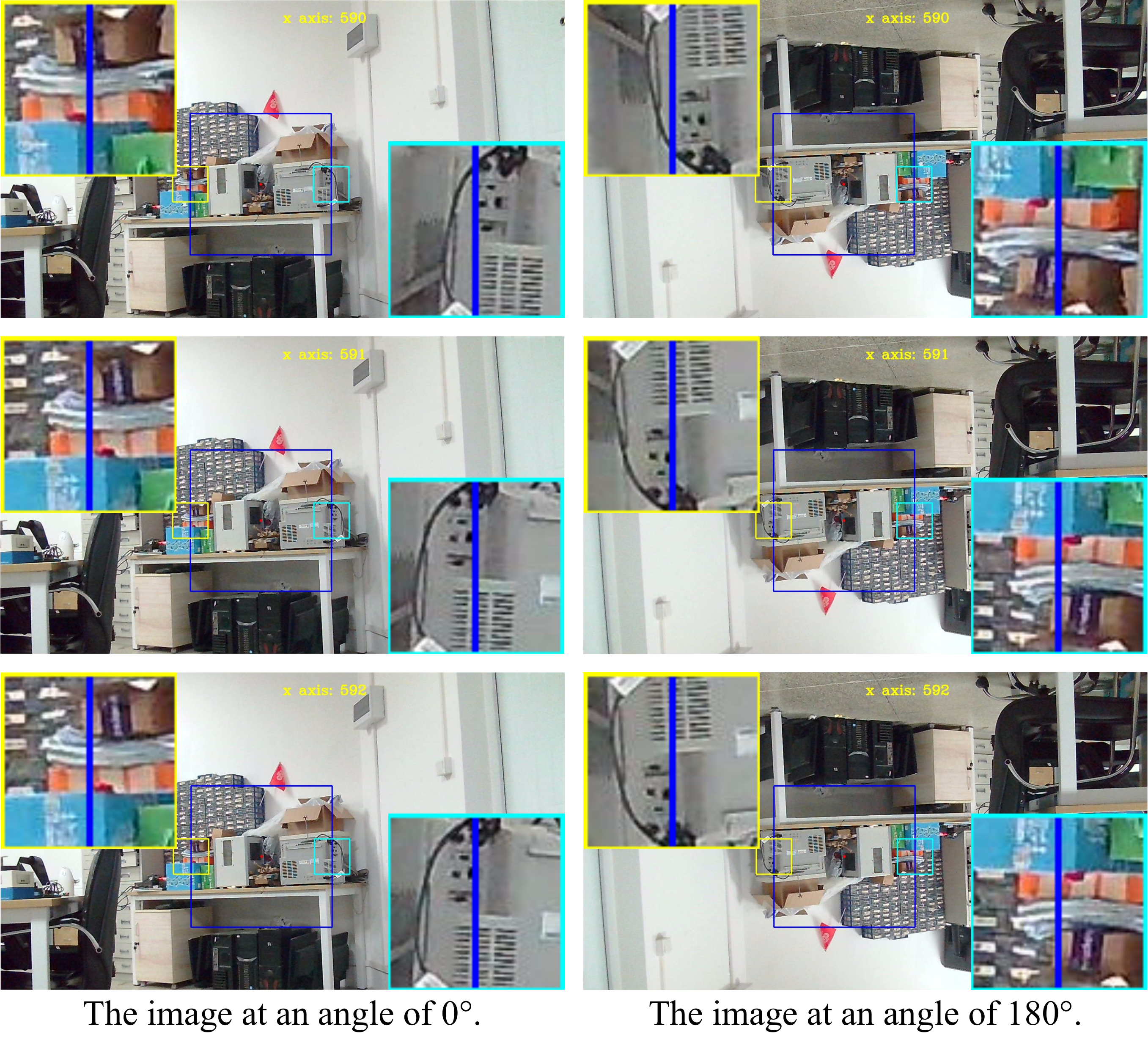}
		\label{Fig:x_aixs}}
	\caption{A real-world example of rotation center identification.}
	\label{Fig:center_cali}
\end{figure*}

Fig. \ref{Fig:y_aixs} illustrates the y-axis identification process. The y-axis coordinate of the red point (the candidate rotation center in the first step of Section \ref{Sec:Identification}) is marked in the top center position above each image. The tangential areas located at the top and bottom of the blue reference box are enlarged and placed in yellow and cyan boxes, respectively. Simultaneously, to facilitate the observation, the image in the cyan box is vertically flipped. As for the object in the second step of Section \ref{Sec:Identification}, we select the 'red cap'. At an angle of 0°, we carefully adjust the position of the rotating platform under different y-axis so that the red cap is tangent to the blue reference box along the y-axis. See the yellow boxes in the first column of Fig. \ref{Fig:y_aixs}. Next, following the third step in Section \ref{Sec:Identification}, the camera is rotated by 180°. When the y-axis coordinate of the red point is 416 or 418, the rotated red cap is clearly not tangent to the blue reference box. See the cyan boxes in the first and third rows of the second column in Fig. \ref{Fig:y_aixs}. Therefore, 416 or 418 is not the rotation center along the y-axis. When the y-axis coordinate of the red point is 417, the rotated red cap is still tangent to the blue reference box, as shown in the cyan box in the second row of the second column. Therefore, we consider 417 as the rotation center along the y-axis.

The x-axis identification process is similar to the y-axis one. Fig. \ref{Fig:x_aixs} shows the results. The tangential areas on the left and right sides of the blue reference box are zoomed in and placed in yellow and cyan boxes, respectively. For better observation, the image in the cyan box is horizontally flipped. At an angle of 0°, the rotating platform is carefully adjusted to make the red cap tangent to the blue reference box. The yellow boxes in the first column of Fig. \ref{Fig:y_aixs} show tangential areas. At an angle of 180°, when the x-axis coordinate of the red point is 591, the rotated red cap is still tangent to the reference box (see the cyan box in the second row, second column of Fig. \ref{Fig:x_aixs}), indicating that 591 is the rotation center along the x-axis.

Consequently, we think that the rotation center of the constructed RMB imaging system is (591, 417).

\subsection{The Results of the Rotation Center Error}

\begin{figure*}
	\centering
	\includegraphics[scale=0.19]{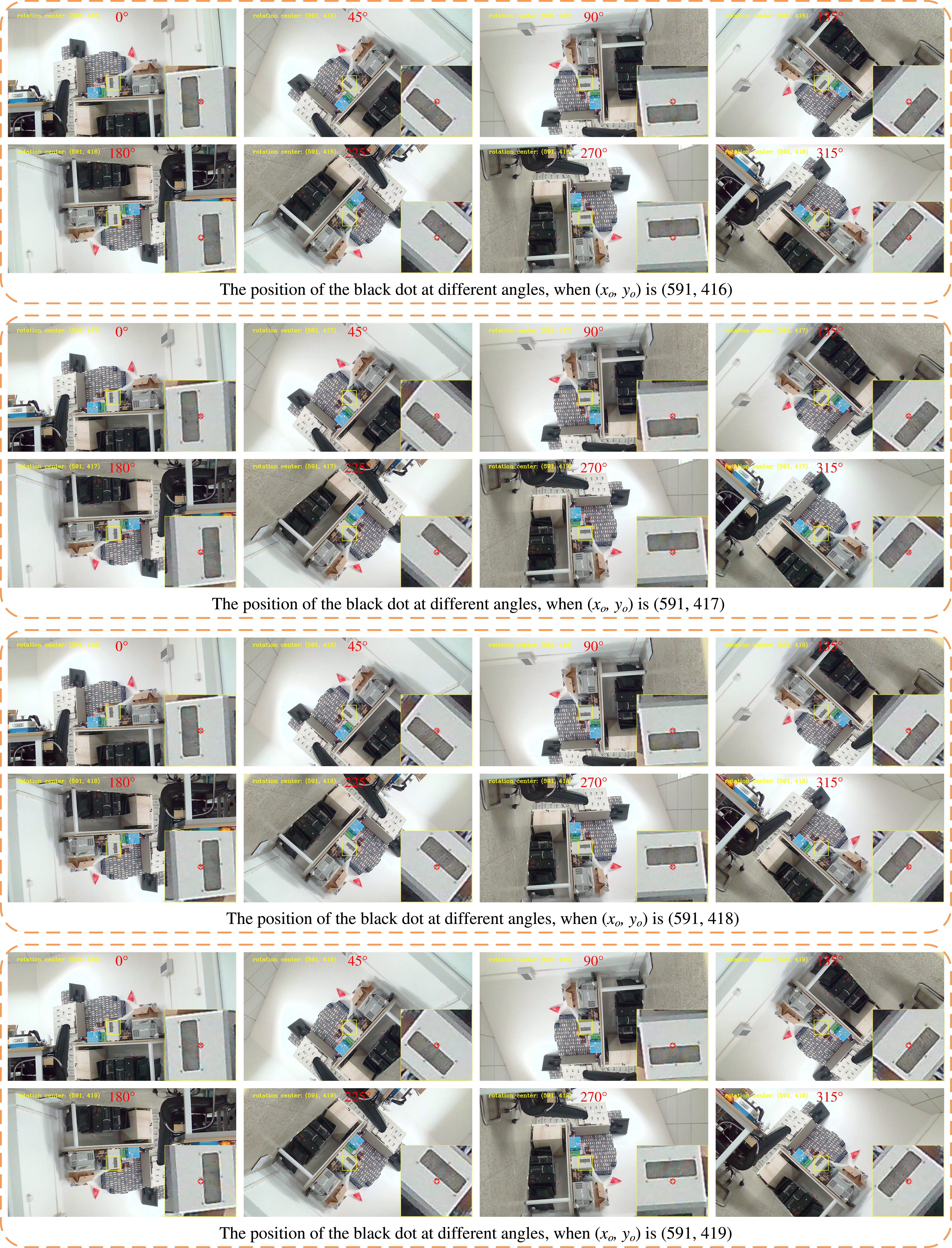}
	\caption{The rotation center error along the y-axis.}
	\label{Fig:center_y}
\end{figure*}

\begin{figure*}
	\centering
	\includegraphics[scale=0.18]{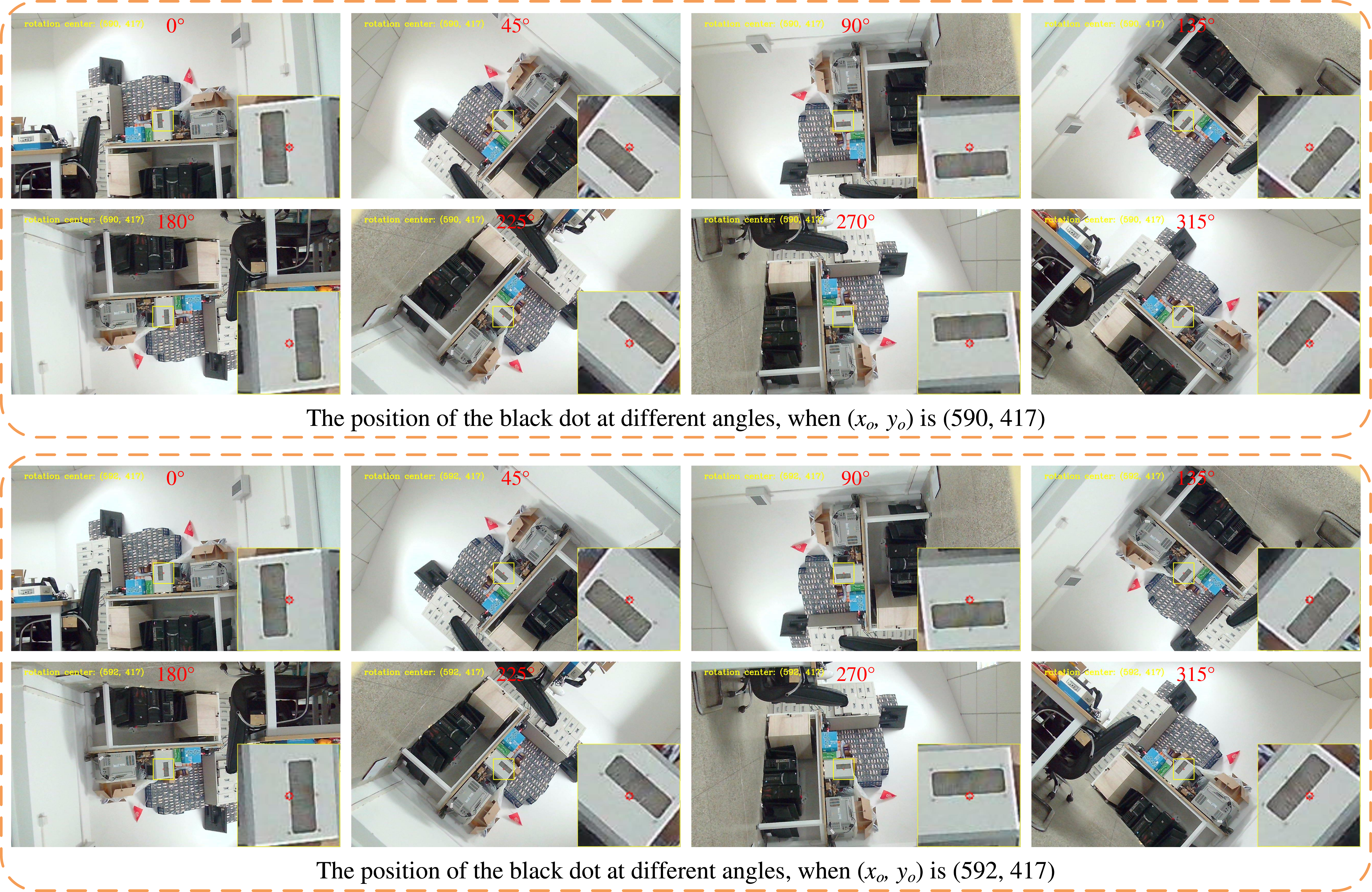}
	\caption{The rotation center error along the x-axis.}
	\label{Fig:center_x}
\end{figure*}

\begin{figure*}
	\centering
	\includegraphics[scale=0.4]{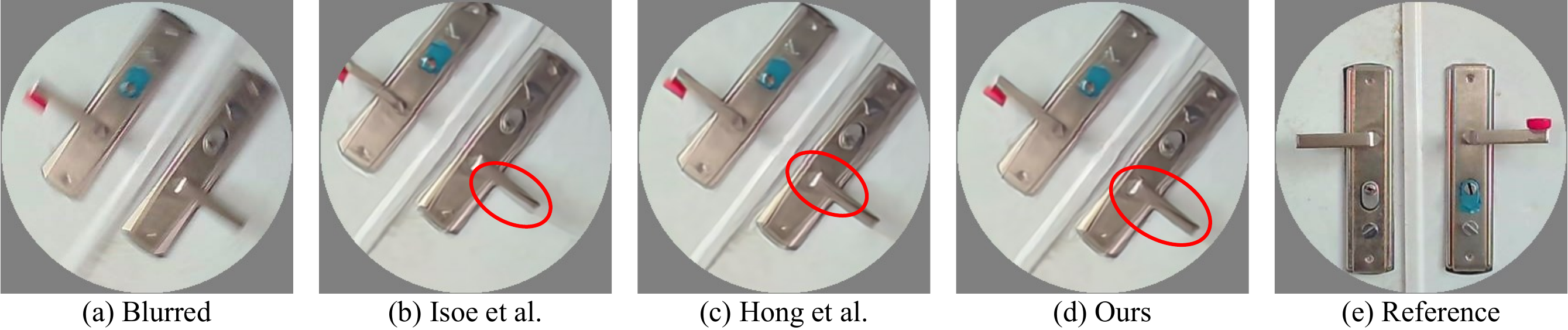}
	\caption{Restored images with different rotation centers.}
	\label{Fig:real_scenes}
\end{figure*}

In this section, we estimate the rotation center error. To demonstrate the error between the identified rotation center $({x}_{o},{y}_{o})$ and the ideal rotation center, we move an object in the field of view to position $({x}_{o},{y}_{o})$ by adjusting the position of the rotating platform. Then, we rotate the rotating platform to different angles. The smaller the change of the object's position in the pixel coordinate system, the smaller the error between $({x}_{o},{y}_{o})$ and the ideal rotation center.

We select a black dot in the scene as the object in this experiment. In Fig. \ref{Fig:center_y} and Fig. \ref{Fig:center_x}, $({x}_{o},{y}_{o})$ is marked in the yellow font at the top left corner in each sub-figure. To facilitate the observation and determination of the black dot's position, we draw a red circle with radius 3, whose center is at $({x}_{o},{y}_{o})$, and its nearby area is enlarged and placed in the yellow box. The current angle of the rotating platform is indicated in red font at the top center. We highly suggest readers watch the videos\footnote{\url{https://pan.baidu.com/s/1bOUjuNUEA3WjQ02aJj7G1Q?pwd=rvzv}} for a better visual effect.

Fig. \ref{Fig:center_y} shows the error analysis of the rotation center along the y-axis. When $({x}_{o},{y}_{o})$ = (591, 416) or (591,419), the black dots significantly deviate from the red circles, indicating larger errors between these points and the ideal rotation center. When $({x}_{o},{y}_{o})$ = (591, 417), the black dot shifts significantly downward, for instance at angles of 180°, 225°, and 270°; When $({x}_{o},{y}_{o})$ = (591, 418), it moves noticeably upward, for example at angles of 135°, 180°, and 225°. This suggests that the ideal rotation center along the y-axis is definitely between 417 and 418. In this paper, 417 is selected as the estimated rotation center along the y-axis. Consider the worst case, where the y-axis ideal rotation is infinitely close to 418, then the error along the y-axis is less than \textbf{1} pixel.

Fig. \ref{Fig:center_x} depicts the error analysis of the rotation center along the x-axis. When $({x}_{o},{y}_{o})$ = (590, 417), the black dots move prominently to the right, for example at angles of 180° and 225°; When $({x}_{o},{y}_{o})$ = (592, 417), the black dot is significantly shifted to the left, for instance at angles of 180° and 225°. This indicates that the ideal rotation center along the x-axis must be between 590 and 592. We choose 591 as the estimated x-axis rotation center in this paper. Considering the worst case, where the ideal rotation center along the x-axis is infinitely close to either 590 or 592, the error would be less than \textbf{1} pixel.

In summary, our experimental results confirm that our identification error of the rotation center along a single axis is less than \textbf{1} pixel. However, according to Section \ref{Sec:Error_Analysis}, this error should be less than 0.5 pixels, which means the black dot moves within a circle with radius = 1. Considering the resonance or vibration of the step motor \cite{vernezi2021vibration,gundogdu2021narma}, the rotation axis of the rotating platform becomes unstable, which is different from the assumptions in Section \ref{Sec:Error_Analysis}. Therefore, the error of the rotation center is bigger than the theoretical one.

\subsection{The effectiveness in RMD}

The purpose of this study is to achieve a more accurate rotation center and enhance the performance of RMD. In this paper, we obtain real RMB images by our constructed RMB imaging system. We rotate the stepper motor at a speed of 3 r/s to capture the RMB image, as shown in Fig. \ref{Fig:real_scenes}(a). Subsequently, a clear image is captured as a reference when the stepper motor is stationary, as depicted in Fig. \ref{Fig:real_scenes}(e).

We estimate the rotation center through Isoe \textit{et al.} \cite{isoe2012blind} and Hong \textit{et al.} \cite{hong2003fast} methods, resulting in (442, 616) and (425, 593) respectively. These estimated rotation centers were used as inputs for Qin \textit{et al.} \cite{qin2023progressive} method to obtain the restored image. It can be observed that due to significant errors in the algorithm-based estimations, namely Isoe \textit{et al.} \cite{isoe2012blind} and Hong \textit{et al.} \cite{hong2003fast} methods, the restored image exhibits ringing artifacts, marked by red circles in Fig. \ref{Fig:real_scenes}.

\section{Conclusion}\label{Sec:Conclusion}

We propose a geometric-based method for determining the rotation center and analyze its error range. Furthermore, we construct an RMB imaging system. Experimental results demonstrate that our method achieves an error of less than 1 pixel along a single axis (x-axis or y-axis). The estimated rotation center helps non-blind RMD approaches achieve better restoration results. Practical experiments conducted on real RMB images confirm that our proposed approach significantly improves the performance of existing RMD methods.

\bibliographystyle{IEEEtran}
\bibliography{arxiv.bib}

\end{document}